\documentclass[runningheads]{llncs}
\usepackage[T1]{fontenc}
\usepackage{amsmath}
\usepackage{amssymb}
\usepackage{xcolor}
\usepackage{graphicx}
\usepackage{subcaption}
\usepackage{hyperref}
\newcommand{\etal}{\textit{et~al.}}

\usepackage[font=small]{caption}

\begin{document}

\title{Interpreting Context-Aware Human Preferences for Multi-Objective Robot Navigation}
%
%
\author{Tharun Sethuraman \inst{1} \and
Subham Agrawal\inst{2, 3} \and
Nils Dengler \inst{2, 3} \and
Jorge de Heuvel  \inst{2}
\and Teena Hassan  \inst{1}
\and Maren Bennewitz  \inst{2, 3}
}

\authorrunning{T. Sethuraman et al.}
%
\institute{Hochschule Bonn-Rhein-Sieg, Germany \and
University of Bonn, Germany \and
Lamarr Institute for Machine Learning and Artificial Intelligence, Germany}

\maketitle              
\begin{abstract}
Robots operating in human-shared environments must not only achieve task-level navigation objectives such as safety and efficiency, but also adapt their behavior to human preferences. 
However, as human preferences are typically expressed in natural language and depend on environmental context, it is difficult to directly integrate them into low-level robot control policies. In this work, we present a pipeline that enables robots to understand and apply context-dependent navigation preferences by combining foundational models with a Multi-Objective Reinforcement Learning (MORL) navigation policy. 
Thus, our approach integrates high-level semantic reasoning with low-level motion control. 
A Vision-Language Model (VLM) extracts structured environmental context from onboard visual observations, while Large Language Models (LLM) convert natural language user feedback into interpretable, context-dependent behavioral rules stored in a persistent but updatable rule memory. A preference translation module then maps contextual information and stored rules into numerical preference vectors that parameterize a pretrained MORL policy for real-time navigation adaptation. We evaluate the proposed framework through quantitative component-level evaluations, a user study, and real-world robot deployments in various indoor environments. Our results demonstrate that the system reliably captures user intent, generates consistent preference vectors, and enables controllable behavior adaptation across diverse contexts. Overall, the proposed pipeline improves the adaptability, transparency, and usability of robots operating in shared human environments, while maintaining safe and responsive real-time control.

\keywords{Preference Learning  \and Foundation Models \and Multi-Objective Robot Learning}
\end{abstract}

\section{Introduction}



Robots are increasingly being deployed in environments where they share physical space with humans and are expected to support humans on everyday tasks.
In such human-centered environments, robot behavior is not evaluated solely on task performance but also on compliance with social norms and individual human preferences~\cite{llm-personalize-han-2025}. 
Respecting these norms is essential to ensure user comfort, perceived safety, and long-term acceptance of robotic systems in daily life~\cite{SRLM-wang-2024}. 
However, human preferences on robot behavior are inherently dynamic and highly context-dependent~\cite{human-centered-robots-doncieux-2022}. 
For example, preferred robot navigation behavior may change depending on context such as room type (e.g., kitchen versus bedroom), objects in the vicinity, human activity, or social situation. 
Therefore, robots must be capable of contextual reasoning and adaptation in order to align their behavior with human expectations. 
Recent advances in foundational models have enabled the development of general-purpose reasoning systems with strong generalization ability and contextual understanding~\mbox{\cite{llm-survey-kim-2024,social-nav-vlm-song-2025}}. 
These systems demonstrate the ability to extract semantic information, infer human intent~\cite{mannering2024generative}, and reason over complex multimodal observations~\cite{suzuki2022survey}, making them promising tools to enable more adaptive and socially-aware robotic behavior.


\begin{figure}[t]
    \centering
    \includegraphics[width=\linewidth, trim= 100 0 100 0, clip]{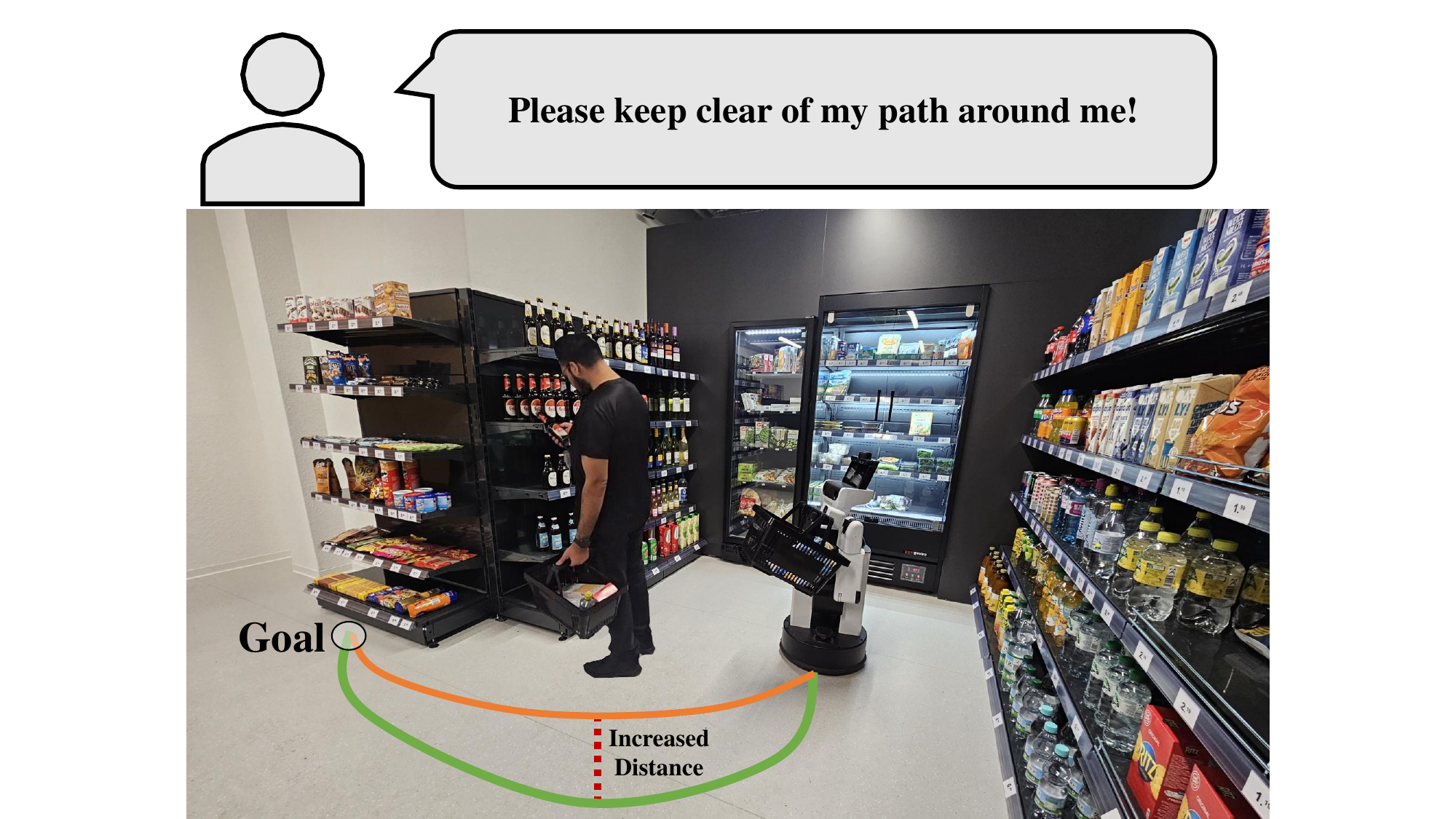}
    \caption{Preference-conditioned navigation behavior in a human-shared supermarket environment.
Given a natural language user instruction requesting increased personal clearance, the robot adapts its trajectory to maintain a larger distance from nearby humans while progressing toward the goal location. As can be seen from the baseline shortest-path trajectory~(orange) and the preference-adapted trajectory~(green), the expressed preference influences the navigation behavior.}
    \label{fig:Cover_of_proposed_approach}
\end{figure}

Despite their strong reasoning capabilities, directly integrating LLMs/VLMs into real-time robot control pipelines remains challenging. 
Foundational models typically introduce high inference latency and require substantial computational resources, which conflicts with the real-time constraints of robot motion control. 
These models often suffer from hallucinations, producing outputs that appear plausible but are factually incorrect, physically infeasible, or do not align with the agent's actual state and environment.
This could lead to unpredictable and unsafe behavior if used directly for closed-loop control.
Therefore, while these models can provide high-level semantic reasoning and decision support, they are impractical for direct low-level control in safety-critical and latency-sensitive robotic applications.

To address this limitation, hybrid control architectures have emerged as a promising solution. 
In particular, low-latency motion policies such as Multi-Objective Reinforcement Learning (MORL) provide an effective mechanism for real-time control while supporting multiple competing objectives~\mbox{\cite{kawaharazuka_realworld_2024,morl-jorge-2025}}. 
In MORL-based navigation, robot behavior is modulated through a numerical preference vector that dynamically weights objectives such as efficiency, velocity, or human comfort. 
This formulation enables flexible runtime adaptation of navigation behavior and allows the robot to balance multiple objectives in response to changing environmental conditions.
However, although MORL policies enable efficient behavior adaptation, its numerical preference vectors are difficult to intuitively specify and interpret for human users, creating a misalignment between high-level human intent and low-level control parameterization.
This can reduce user trust and hinder practical deployment in real-world  scenarios.

In this work, we propose a preliminary framework that enables robots to \emph{elicit}, \emph{represent}, and \emph{apply} context-dependent human navigation preferences using foundational models such as Large Language Models (LLMs) and Vision Language Models (VLMs), while preserving low-latency control through a MORL-based navigation policy. 
Rather than using foundational models for closed-loop control, we restrict them to high-level reasoning to extract the semantic context from perception, interpret user feedback, and convert it into structured parameters that condition a pretrained MORL policy at runtime (see Fig.~\ref{fig:Cover_of_proposed_approach}).

Specifically, our approach decomposes preference understanding into three stages.
First, a VLM extracts a structured description of the current scene (e.g., room type, relevant objects, human presence, and lighting), which serves as contextual grounding. 
Then, an LLM  encodes user preferences expressed in natural language by converting them into human-readable, context-dependent rules that are stored in a rule memory, which can be updated during runtime, for later retrieval. 
Finally, a preference translation module maps the context and the retrieved rules for the current situation to a numerical preference vector using a second LLM. The vector parameterizes the MORL policy, enabling the robot to adapt its navigation behavior in real time without retraining.

By combining the semantic reasoning and language grounding of LLMs/VLMs with the execution efficiency of MORL, our framework enables intuitive preference specification and human understandable preference-conditioned robot navigation, while maintaining responsive low-level control. Additionally, our human understandable and editable ruleset allows transparent and analyzable human preference as opposed to black box nature associated with foundational models.
In summary, the main contributions of our work are:
\begin{itemize}
    \item A pipeline for preference-aware navigation that (i) predicts scene context from onboard vision, (ii) elicits, updates, and maintains context-dependent user preferences as an interpretable rule memory, and (iii) translates these rules into MORL preference vectors for low-latency motion control.
    \item An experimental evaluation of individual system components on standard datasets, a user study, and real-robot deployments in various environments, demonstrating that the resulting navigation behavior adapts to context-dependent user preferences.
\end{itemize}
The codebase and the prompts used for each component can be found on \href{https://github.com/HumanoidsBonn/Interpreting-Context-Aware-Human-Preferences-for-MORL}{github}.

\section{Related Work}

\subsection{Context-Aware Social Navigation}
Enabling robots to navigate effectively in shared environments is a highly researched task and requires a grounded understanding of both environmental and social contexts~\cite{mavrogiannis2023core}~\cite{ngo2022socially}~\cite{mustafa2024context}. 
Early approaches in the domain of context-aware navigation used hand-crafted rules and defined static and dynamic contexts~\mbox{\cite{haarslev2021context,cosgun2017context}}. 
While static contexts, such as the room type, are not expected to change during navigation, dynamic contexts evolve over time and include factors such as changing navigation objectives and the density of humans in the environment. More recently,
Stefanini~\etal~\cite{stefanini2024efficient} showed the effectiveness of using multiple contextual cues from humans, such as 2D position, 3D body pose, 2D velocity, and activity~(walking, standing) to enable smooth navigation in shared environments. 
While effective, these approaches are limited in the scope of contextual understanding and  do not account for other relevant aspects of the scene outside humans and handcrafted rule sets. 

Jia~\etal~\cite{jia2023deep} proposed a learning-based approach that utilizes neural networks for object detection to recognize hazardous objects (e.g., elevators, glass doors, staircases) in indoor environments for safe navigation. Benisetty~\etal~\cite{banisetty2021deep} combine visual context classification and spatial human-configuration analysis to autonomously select socially appropriate navigation objectives across multiple interaction scenarios. However, these approaches are typically bound to their training data and do not generalize to other objects or contexts beyond.
Here, current advances in foundational models can be used to help bridge this gap. 
The zero-shot capability of VLMs~\cite{han2025multimodal} enables robots to acquire environmental context without extensive task-specific training~\cite{sathyamoorthy2024convoi,narasimhan2025olivia}.
In particular, Sathyamoorthy~\etal~\cite{sathyamoorthy2024convoi} use a VLM to classify the scene into differnt environmental types, such as indoor corridor or outdoor terrain, to directly inform the robot's navigation strategy. 
Similarly, Narasimhan~\etal~\cite{narasimhan2025olivia} utilize a VLM to distill social context generation into a more lightweight model to support faster inference and lifelong learning.
Our work leverages this capability and extends it by also extracting and considering information about other aspects such as lighting conditions and relevant objects in the scene influencing robot navigation.

\subsection{Reflecting User Preference in Social Navigation}
Early work on preference-aware robot navigation primarily relied on iterative user interaction to adapt trajectories in known environments~\cite{wilde2018learning} or on model-based reinforcement learning to personalize navigation behavior over time~\cite{ohnbar2018personalized}. 
While these approaches demonstrated improved personalization, they require continuous data collection and repeated online adaptation, which can limit scalability and deployment in real-world scenarios.
To reduce this dependency, offline reward learning methods were introduced to capture human navigation preferences directly from recorded data~\cite{seneviratne2025halo}.
These approaches improve generalization to unseen environments, but they remain limited when user preferences change, since the reward model must typically be retrained to reflect new behaviors.
Therefore, de Heuvel~\etal~\cite{morl-jorge-2025} proposed to combine demonstration learning with multi-objective reinforcement learning to enable post-training adaptation to changing user preferences without retraining, by modulating the influence of demonstrations and further navigation objectives during deployment.

Recently, foundational models have enabled more flexible and scalable personalization~\cite{mahadevan2024generative,wu2023tidybot}. 
For example, LLMs have been used to jointly reason over perception, navigation, and user interaction, allowing robots to interpret natural language instructions and navigate toward user-specified goals~\cite{dai2024think}. 
Similarly, our work leverages LLMs and VLMs to translate natural language elicited user preferences into context-dependent behavioral rules that adapt to the environment.

Beyond language understanding, VLA models enable direct mapping from multimodal observations to low-level robot actions~\cite{kawaharazuka2025vla-survey}. 
However, directly executing foundational model outputs can introduce safety risks and reduce interpretability due to the black box nature of their decisions. 
To address this, hybrid approaches combine foundational high-level reasoning with conventional low-level control architectures.
For example, Hwang et al.~\cite{hwang2024promptable} propose a multi-objective reinforcement learning framework that personalizes agent behavior by inferring reward weights from human input.
Although this allows learning behavior based on human preferences, their approach considers static preferences represented as a fixed weighted sum of objectives, while ignoring contextual information. 
Martinez~\etal~\cite{martinez2025hey} improve upon this and use language instructions along with optional image inputs to adjust the cost terms within an MPC controller.
In contrast, our framework maintains a persistent, context-aware rule set and integrates real-time vision-language grounding together with natural language explanations to improve transparency and user trust.
Our work also allows dynamic adjustment of robot behavior as the environment changes due to constant context updates received from the context predictor VLM and the preference translator using the context with the existing rule set to modify the navigation strategy.
\newpage

\newpage

\section{Problem Definition}
\label{sec:PD}
To enable adaptive navigation under changing user preferences, we formulate navigation as a multi-objective reinforcement learning (MORL) problem~\cite{morl-jorge-2025}. In MORL, the agent optimizes a vector-valued reward function
\begin{equation}
\mathbf{r}_t = (r_t^{(1)}, \dots, r_t^{(N)}) \in \mathbb{R}^N,
\end{equation}
where each component corresponds to a navigation objective. The number of objectives $N$ denotes the dimensionality of the reward vector and depends on the considered task. In our setting, these objectives include core navigation requirements, i.e., collision avoidance and goal reaching, and preference-dependent behavioral objectives such as efficiency, human and object proximity, and robot velocity.
In the MORL formulation, preferences are represented by a preference weight vector $\boldsymbol{\lambda}_t \in [0,1]^N$, which parameterizes a reward function
\begin{equation}
\boldsymbol{\lambda}_t^\top \mathbf{r}(o_t, a_t).
 \end{equation}
This formulation enables a single navigation policy to adapt its behavior across the preference space by conditioning on $\boldsymbol{\lambda}_t$.
Therefore, $\boldsymbol{\lambda}_t$ controls the trade-off between different navigation objectives by weighting their contribution to the policy optimization and action selection. As demonstrated in preference-driven MORL navigation~\cite{morl-jorge-2025}, conditioning the policy on the preference vector allows continuous interpolation between navigation styles without retraining. 

In this work, we consider a mobile robot operating in a partially observable environment shared with humans and objects.
At each time step $t$, the robot receives multimodal observations 
$\mathbf{o_t}$, including RGB visual input from onboard cameras and LiDAR data to support robust navigation. In addition, the robot receives human preference feedback $f_t$ expressed in natural language, describing desired behavioral adjustments. 
We therefore define a preference-conditioned navigation policy
\begin{equation}
\pi(a_t \mid \boldsymbol{o_t}, \boldsymbol{b_t}, \boldsymbol{\lambda}_t),
\end{equation}
that produces the pair of translational and rotational velocities $a_t$, consistent with both general navigation task objectives $\boldsymbol{b}_t$, i.e., collision-free goal reaching, and user-specific behavioral preferences encoded in $\boldsymbol{\lambda}_t$.
The key challenge is that user preferences~$f_t$ are not directly expressed in the structured form required by the downstream navigation policy. Instead, they must be grounded in the current environmental context $\mathcal{C}_t$, translated into interpretable behavioral rules, and transformed into structured parameters suitable for the multi-objective policy execution. To address this challenge, we formulate preference alignment as the task of generating the context-conditioned preference vector
\begin{equation}
\boldsymbol{\lambda}_t = \Phi(\boldsymbol{o}_t^{\mathit{RGB}}, f_t),
\end{equation}
where $\Phi(\cdot)$ denotes a multimodal reasoning pipeline that integrates visual scene understanding and natural language preference interpretation.
In the following, we present the individual components of $\Phi(\cdot)$, namely context prediction, rule updating, and preference translation. Together, these components generate the preference vector $\boldsymbol{\lambda}_t$, which parameterizes the MORL navigation policy and enables adaptive, preference-aligned robot behavior.

\section{Our Approach}

We propose a multimodal reasoning pipeline for understanding context-based user preferences through foundational models and adapting a navigation policy to reflect these expressed preferences in robot behavior during runtime.

\subsection{Overview}

Our pipeline leverages three complementary foundational models to enable reliable translation of user preferences into navigation objectives.
First, a \textbf{context predictor V} identifies environmental features using a VLM, which extracts contextual information from RGB images streamed from the robot’s camera: 
\begin{equation}
        C_t = V(o_t^{\text{RGB}})
\end{equation}

Then, a \textbf{rule updater U} converts human feedback~$f_t$ into human-readable, context-dependent rules by translating user preferences into a set of general behavioral guidelines for the robot using an LLM. The main task of the rule updater is to update the previous rule set ($R_{t-1}$) with the new one ($R_{t}$) after considering the context and the human preferences: 
\begin{equation}
        \mathcal{R}_{t} = U(f_t, C_t, \mathcal{R}_{t-1})
\end{equation}

Finally, a \textbf{preference translator P} combines contextual information and extracted rules using another LLM to generate a preference vector for the MORL navigation policy~\cite{morl-jorge-2025}:
\begin{equation}
        \lambda_t = P(C_t, \mathcal{R}_{t})
\end{equation}

\begin{figure}[t]
    \centering
    \includegraphics[width=\linewidth]{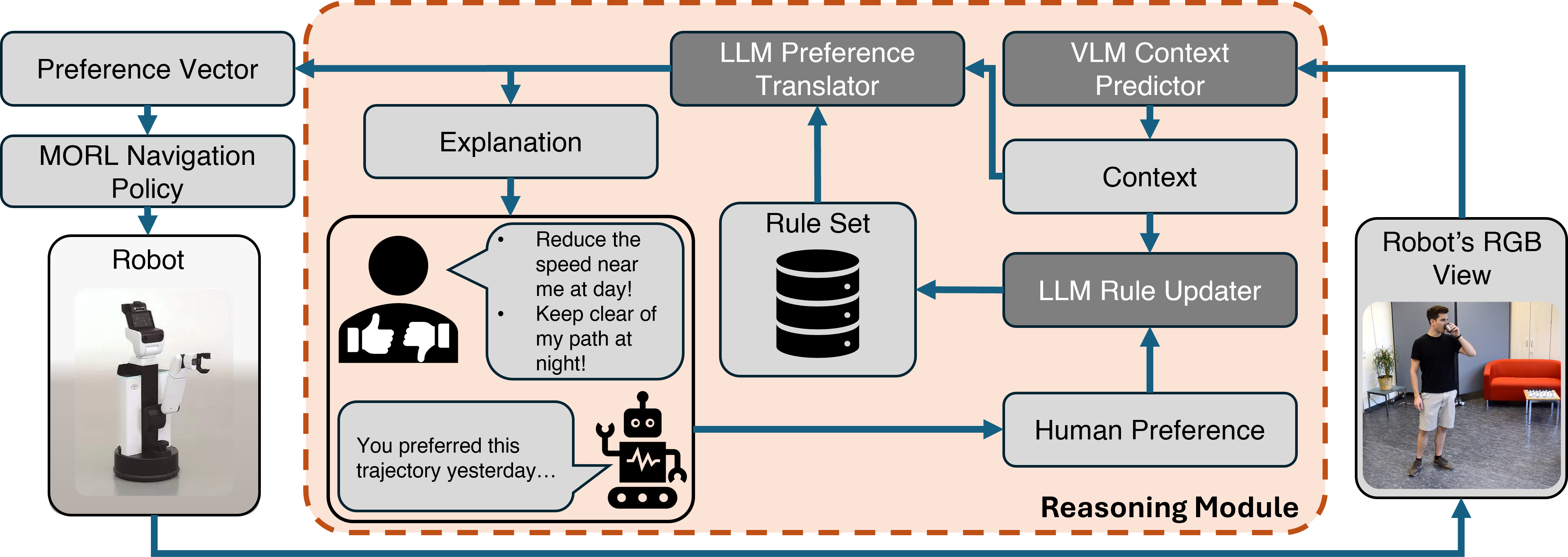}
    \caption{Overview of the proposed preference-aware navigation architecture.
The reasoning module integrates VLM-based \textbf{context prediction}, LLM-based \textbf{rule updating}, and \textbf{preference translation} to convert visual context and natural language user feedback into structured rules and preference vectors usable by a MORL agent for online low-level navigation control. The preference vectors condition the navigation policy, enabling the robot to adapt its behavior to context-dependent user preferences while providing interpretable explanations of its decisions.}
    \label{fig:Architecture_of_proposed_approach}
    \vspace{-10px}
\end{figure}

An overview of our proposed approach is illustrated in Fig.~\ref{fig:Architecture_of_proposed_approach}. The individual components are described in detail in the following.

\subsection{Context Predictor}
\label{sec:CP}
\begin{figure}[t]
    \centering
    \includegraphics[width=\linewidth]{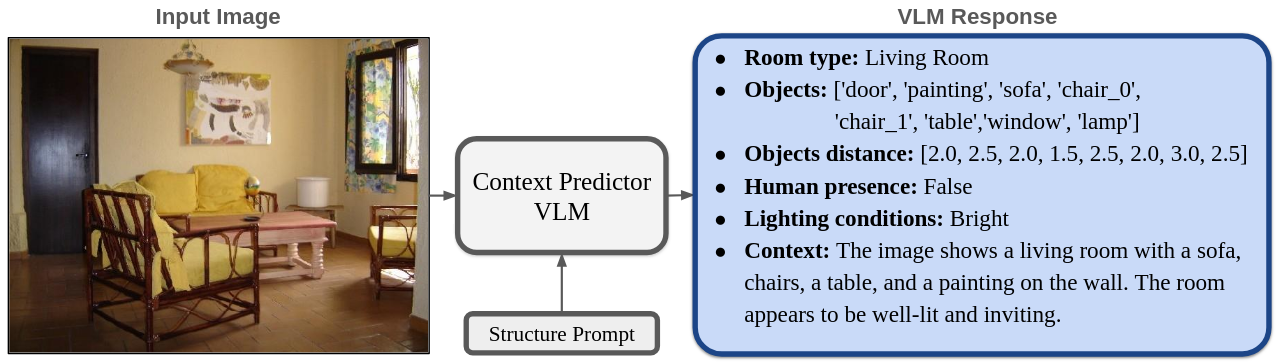}
    \caption{
    Example output of the context predictor. Given an input RGB scene image and a structured prompt, the VLM-based context predictor extracts semantic and spatial environmental information, including room type, detected objects, approximate object distances, human presence, lighting conditions, and a natural language scene description. This structured context representation is used as grounding input for downstream rule generation and preference translation modules. 
    }
    \label{fig:context}
    \vspace{-10px}
\end{figure}

The context predictor is responsible for extracting structured, semantically meaningful environmental information from raw visual observations. Since user preferences are inherently grounded in the surrounding environment, reliable context estimation is important for translating natural language feedback into navigation behavior. In our pipeline, the context predictor provides a structured representation of the scene that is later used by downstream reasoning modules to ground user preferences and generate navigation constraints.
An example output of our context predictor is illustrated in Fig.~\ref{fig:context}.
At each inference step, an RGB-image, captured from the robot’s onboard camera stream, is provided as input to the context predictor together with a structured instruction prompt. The prompt explicitly defines the contextual attributes that must be extracted, provides semantic definitions for each attribute, and specifies the required output format. 
We implemented the context predictor using a VLM (Gemini 2.0 Flash), which enables joint visual understanding and structured semantic reasoning. 
Compared to traditional perception pipelines based on independent object detection, semantic segmentation, and scene classification modules, the VLM allows unified extraction of both low-level perceptual information and high-level semantic context. This is particularly important for preference grounding, as human preferences are often expressed using abstract or relational descriptions, e.g., “stay further away from people in crowded areas” or “move more slowly in dim lighting”. The contextual attributes extracted by the VLM include:
\begin{itemize}
    \item \textbf{Objects}: Open-vocabulary object detection identifies relevant scene objects, including humans.
    \item \textbf{Distance to Objects}: Estimation of approximate relative distances between the robot and detected objects, providing spatial grounding for rule generation and preference translation.
    \item \textbf{Human Presence}: A binary indicator specifying whether humans are present in the scene, enabling activation of socially-aware behaviors.
    \item \textbf{Lighting Condition}: The global illumination level is classified into a fixed set $\{$Bright, Gentle, Low$\}$ to ensure stable downstream conditioning.
    \item \textbf{Room Type}: The environment is mapped to a predefined semantic category $\{$Kitchen, Living room, Dining room, Bed room$\}$, with fallback to open descriptions if necessary.
    \item \textbf{Scene Context}: A short (1--2 sentence) natural language summary describing the overall scene and activity context.
\end{itemize}
The VLM output is serialized into a structured JSON representation to ensure compatibility with downstream LLM modules.
Therefore, the context predictor transforms high-dimensional visual observations into a structured symbolic representation $\mathcal{C}_t$ that captures both geometric and semantic scene properties.

\subsection{Rule Updater}

The rule updater converts user preferences $f_t$ expressed in natural language into structured, context-dependent behavioral rules that can be stored and  applied during navigation.(Fig.~\ref{fig:rule}). 
We realize this process using an LLM, which enables robust interpretation of textual user feedback. 
By representing rules in natural language to improve system transparency, we allow users to inspect or modify stored preferences.
From a decision-making perspective, preference alignment is treated as a multi-objective problem. The robot must simultaneously satisfy core navigation objectives while adapting behavior based on user preferences. Thus, the rule updater maintains two complementary rule categories:

\begin{figure}[t]
    \centering
    \includegraphics[width=\linewidth]{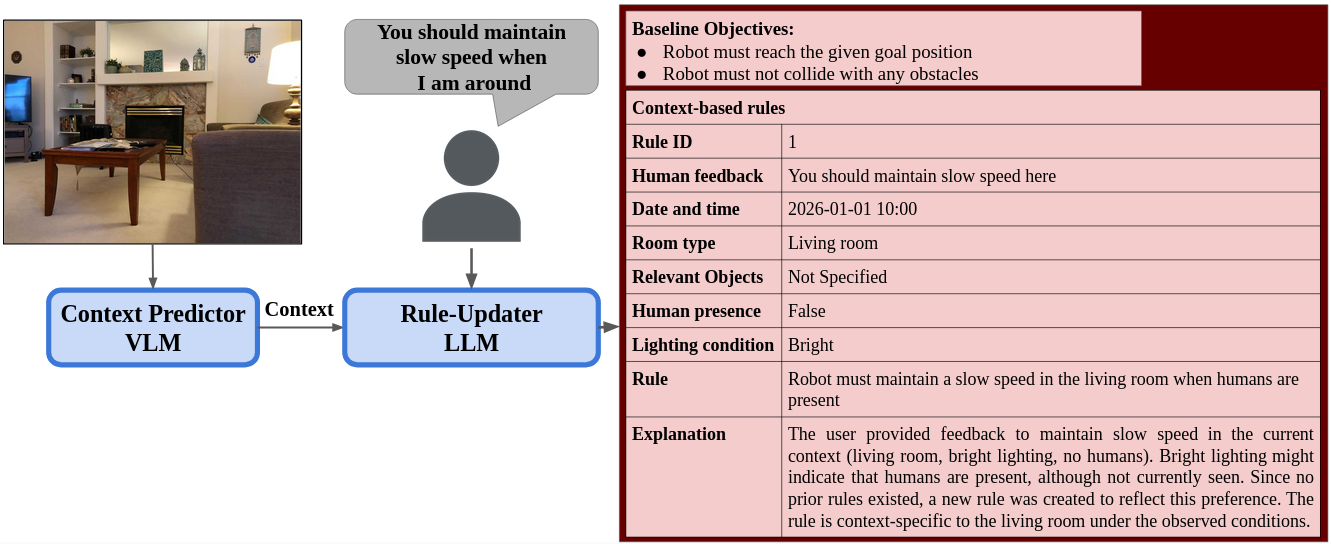}
    \caption{Example rule generation for the context-aware preference pipeline.
    Given an input scene image and a natural language user preference, the VLM-based context predictor extracts structured environmental context, which is combined with user input by the LLM-based rule updater to generate a context-dependent navigation rule and explanation while preserving baseline safety objectives.
    }
    \label{fig:rule}
\end{figure}

\begin{itemize}
    \item \textbf{Baseline Objectives}: These represent fundamental navigation requirements that must always be satisfied, independent of user preferences or context. In this work, baseline objectives include reaching a given goal position and avoiding collisions. These objectives  are treated as non-negotiable constraints (examples are shown in Fig.~\ref{fig:rule}). While the MORL agent was conditioned on these baseline objectives during training, they are still provided to the rule updater to avoid updating conflicting rules into the rule set.
    
    \item \textbf{Context-Based Rules}: These rules encode user preferences that are only applied under specific contextual conditions. Each rule consists of the user's preference together with its corresponding room type, lighting condition, human presence, and relevant scene objects, as discussed in Sec.~\ref{sec:CP}. These rules are stored in a structured database to enable efficient retrieval and modification.
\end{itemize}

For each query, the rule updater receives a structured prompt containing contextual information from the context predictor, the baseline objectives, the existing context-based rule set, and the latest user preference input. The LLM then performs three key reasoning steps. 
First, it extracts relevant contextual cues directly from the user’s natural language feedback. 
Then, comparing with the input rule set, if contextual information is missing or ambiguous, the updater combines it using the output of the context predictor through addition, modification, or deletion. 
Finally, the LLM generates or updates rules in a structured format. 
In total, our rule updater supports three primary rule management operations:

\begin{itemize}
    \item \textbf{Rule Addition}: When a new preference is provided, the LLM evaluates it for potential conflicts with the baseline objectives. If no conflicts are detected, a new context-dependent rule is generated and added to the rule set.
    
    \item \textbf{Rule Modification}: When a user refines an existing preference, the LLM is prompted to identify matching rules and update them while preserving consistency with the baseline objectives. 
    
    \item \textbf{Rule Deletion}: When new preferences contradict existing rules, conflicting rules are removed and replaced with updated rules reflecting the new preference. 
\end{itemize}
Each generated rule is stored together with an explanation describing how the user preference was translated into the rule to improve transparency and user trust by making the reasoning process interpretable.

\subsection{Preference Translator}
Finally, the preference translator is responsible for converting the structured rule set into a numerical preference representation that can directly parameterize a downstream MORL navigation policy. 
The preference translator therefore acts as the connection between context-dependent preference reasoning and continuous motion control.

\begin{figure}[t]
    \centering
    \includegraphics[width=\linewidth]{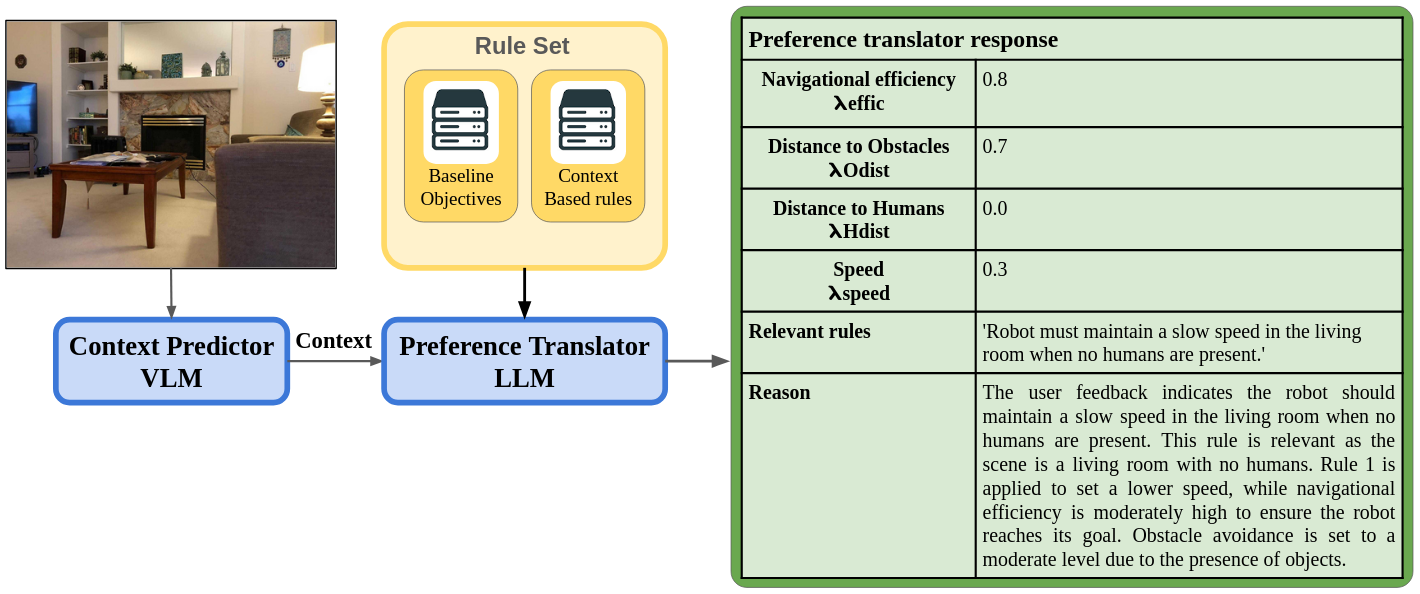}
    \caption{
    Example output of the preference translator. Given the predicted scene context~(living room, no humans present) and the stored rule set, the LLM-based preference translator selects relevant context-based rules and generates a four-dimensional MORL preference vector. The resulting vector modulates navigation efficiency, obstacle distance, human distance, and velocity, while also providing the applied rule and a natural language explanation to ensure transparency in preference-conditioned navigation behavior.
    }
    \label{fig:preferences}
\end{figure}

We implemented the preference translator using an LLM that interprets contextual information and relevant rules and maps them into a continuous preference vector. 
As described in Section~\ref{sec:PD}, the MORL navigation policy represents behavioral trade-offs using a tunable preference vector $\boldsymbol{\lambda}$, where each of the $N$ component corresponds to a specific navigation objective.
In previous work~\cite{morl-jorge-2025}, two baseline objectives were considered: navigational efficiency and distance to humans. 
In this work, the preference vector is altered to include two additional objectives: distance to obstacles and robot velocity so that our preference vector is defined as:
\begin{equation}
\boldsymbol{\lambda} = (\lambda_{\text{effic}}, \lambda_{\text{Odist}}, \lambda_{\text{Hdist}}, \lambda_{\text{velocity}})
\label{equation: preference_vector}
\end{equation}

Each component $\lambda_i \in [0,1]$ determines the weighting of the corresponding navigation objective.
Lower values reduce the influence of an objective, while higher values increase its importance. 
For example, higher $\lambda_{\text{effic}}$ prioritizes shortest-path navigation, while higher $\lambda_{\text{Hdist}}$ increases the distance to humans. 
Similarly, $\lambda_{\text{velocity}}$ controls the roboot's velocity.

During each inference step, the preference translator receives a structured prompt containing contextual information from the context predictor and the updated rule set from the rule updater. 
The prompt defines the semantics of each preference dimension, the allowed numerical range, and the required output format. 
The output is generated in a structured format containing the preference vector, the subset of rules used during computation, and a natural language explanation of the reasoning process.  An example of the preference output is shown in Fig.~\ref{fig:preferences}.
The generation of user preferences into continuous control parameters then allows the MORL policy to dynamically adapt navigation behavior according to environmental context and changing user preferences.

\section{Experimental Evaluation}
To evaluate our proposed pipeline, we conducted a set of experiments targeting both, individual components and the overall system behavior. 
The evaluation consists of three parts: (i) quantitative component-level evaluations of the context predictor and preference translator, (ii) a user study comparing multiple LLMs for rule generation and preference translation, and (iii) real-world robot experiments conducted at three distinct locations to demonstrate system robustness and generalizability.

\subsection{Quantitative Evaluation}
\label{sec:quant}
\begin{table}[t!]
\centering

\begin{tabular}{lccc}
\textbf{Model} & \shortstack{\textbf{Room Type}\\\textbf{Acc. (\%)}} & \shortstack{\textbf{Object}\\\textbf{Acc. (\%)}}
& \shortstack{\textbf{Mean Inf.}\\\textbf{Time (s)}} \\

\hline
\textbf{Gemini-2.0-flash} & \textbf{98.6} & 53.6 & \textbf{2.2} \\
Gemini-1.5-flash & 98.6 & 48.0 & 2.9 \\
Pixtral-large-2411&98.6 & \textbf{54.2} & 9.2 \\
Pixtral-12B      & 97.1 & 47.2 & 4.2 \\
GPT-4o-Mini      & 98.6 & 42.2 & 5.0 \\
GPT-4o           & 98.6 & 44.1 & 7.0 \\
\hline
\end{tabular}%
\caption{Performance comparison of VLMs for context prediction. 
}
\label{tab:vlm_performance}
\end{table}
\paragraph{\textbf{Context Predictor Evaluation:}}
We first perform a comparative evaluation of six candidate VLMs (Table~\ref{tab:vlm_performance}) to identify the most suitable model for context prediction. 
The models were evaluated on 70 images from the MIT Indoor Scenes dataset~\cite{MIT-indoor-scenes-quattoni-2009} using three metrics: room-type classification accuracy, object recognition accuracy, and inference time across all evaluation images. 
Object recognition accuracy was computed as the ratio between correctly detected objects and the total number of ground-truth objects. 
Please note that, due to the rapid development of foundational and reasoning models, the presented evaluation represents a snapshot of model performance at the time of writing.

While most models achieved high room classification accuracy ($>97\%$), \textbf{Gemini-2.0-flash} provided the best overall trade-off between perception performance and latency, achieving the second-highest object recognition accuracy while maintaining the lowest mean inference time, which is important for online deplyment. 

\begin{table}[t]
\centering
\setlength{\tabcolsep}{3pt}

\begin{tabular}{l|cc|cc|cc|cc}
\textbf{Pref. Obj.} &
\multicolumn{2}{c|}{\textbf{Mistral L. 2.1}} &
\multicolumn{2}{c|}{\textbf{GPT-4o}} &
\multicolumn{2}{c|}{\textbf{GPT-4o-mini}} &
\multicolumn{2}{c}{\textbf{Gemini-2.0-fl}} \\
\cline{2-9}
& \textbf{cts} & \textbf{disc}
& \textbf{cts} & \textbf{disc}
& \textbf{cts} & \textbf{disc}
& \textbf{cts} & \textbf{disc} \\
\hline
$\lambda_\mathit{effic}$   & \textbf{0.14} & 0.27 & 0.22 & 0.28 & 0.35 & 0.34 & 0.38 & -- \\
$\lambda_\mathit{Odist}$   & 0.14 & \textbf{0.12} & \textbf{0.12} & 0.15 & 0.21 & 0.15 & 0.22 & -- \\
$\lambda_\mathit{Ddist}$   & \textbf{0.04} & \textbf{0.04} & 0.07 & \textbf{0.04} & 0.15 & 0.61 & 0.81 & -- \\
$\lambda_\mathit{velocity}$& 0.12 & \textbf{0.11} & 0.23 & 0.22 & 0.26 & 0.29 & 0.20 & -- \\
\hline
\textbf{Mean} & \textbf{0.10} & 0.12 & 0.14 & 0.15 & 0.19 & 0.29 & 0.44 & -- \\
\hline
\end{tabular}
\caption{Preference prediction error across evaluated LLMs for both continuous~(cts) and discretized~(disc) preference vector outputs.}
\label{tab:expt_overall_errors}
\vspace{-20px}
\end{table}

\paragraph{\textbf{Preference Translator Evaluation:}}
We evaluated candidate LLMs based on their ability to generate MORL preference vectors from natural language instructions. 
Using 27 scenes from the SRIN dataset~\cite{SRIN-othman-2020} paired with handcrafted ground-truth preference vectors $\boldsymbol{\lambda}^{gt}$, we measured the mean prediction error~($E_{pref}$) across four LLMs candidates: GPT-4o-mini, GPT-4o, Mistral~L.~2.1, and Gemini-2.0-flash. $\boldsymbol{\lambda}^{gt}$ was constructed to reflect minimal (0), medium (0.5), and maximal (1) objective weights for preference values.
For this experiment, we used the full pipeline, including context prediction and rule updating.
To ensure realistic scene and preference inputs, we sequentially provided the pipeline with images and associated preferences collected during a user study (see Sec.~\ref{sec:US}) to construct the rule set.
The preference translator then generated preference vectors conditioned on both the extracted contextual information and the resulting rule set.
We evaluated both discrete and continuous output domains, where the discrete domain is defined over  $\mathcal{D} = \{0.0, 0.1, \dots, 0.9, 1.0\}$, while the continuous domain represents values within $\mathcal{I} = [0,1]$.
The final preference vector prediction performance was measured using mean preference error:

\begin{equation}
E_{pref} = \frac{1}{N} \sum_{i=1}^{N} \big\| \boldsymbol{\lambda}_i^{pred} - \boldsymbol{\lambda}_i^{gt} \big\|.
\end{equation}

As shown in Tab.~\ref{tab:expt_overall_errors}, a statistical analysis using the Wilcoxon signed-rank test highlighted that model selection had a significantly stronger impact on accuracy than the choice of output domain. 
\textbf{Mistral-Large-2.1} achieved the lowest mean error over all preference vector components in the continuous domain, significantly outperforming other candidate models.
These results demonstrate that, although LLMs are not explicitly designed for numerical prediction, the proposed pipeline reliably generates preference vectors suitable for multi-objective robot control.

\subsection{User Study}
\label{sec:US}
To assess whether the proposed pipeline effectively captures and represents natural language user preferences as preference vectors, we conducted a controlled user study. 
The study specifically evaluated the performance and perceived quality of the \textbf{rule updater} and \textbf{preference translator} modules from a human-centered perspective.
Therfore, a within-subjects experimental design was employed with $N = 24$ participants (10 male, 14 female) with a mean age of~29.45~($\pm 5.48$). Two LLM candidates were compared: GPT-4o and Mistral-Large-2.1.

The experimental procedure consisted of three phases: (i) user preference acquisition across different indoor scenes, (ii) evaluation of the rule updater, and (iii) evaluation of the preference translator. At the start of the user study, the participants were asked to fill up a demographics form.
This was followed by the preference acquisition phase, where the participants were shown five images from the SRIN dataset. 
The first image was a trial image and was identical for all participants, while the remaining four images were presented in randomized order to eliminate order bias from the results. 
The participants were instructed to provide their preferences in terms of navigational efficiency (short or long route), distance from humans and obstacles, and robot speed. 
For each image, participants provided natural language preferences which were recorded using a GUI based interface.
In the second phase, the collected preferences were processed by both LLMs to generate context-dependent rules along with natural language explanations. 
Each participant was shown the outputs of both models side-by-side in randomized order and was instructed to evaluate them using a seven point Likert-scale questionnaire.
In the final phase, the preference translator generated preference vectors and corresponding natural language explanations based on the previously generated contextual information and rule sets. 
Participants were asked to compare responses from both models side-by-side and rate the relevance of the extracted rules as well as the clarity of the explanations. 
To reduce presentation bias, extracted rules were shown before detailed explanations in both evaluation phases.

Results show both candidate LLMs received overall positive ratings from participants. For the \textbf{rule updater}, Mistral-Large-2.1 significantly outperformed GPT-4o in generating interpretable rules ($\textbf{6.35} \pm \textbf{1.04}$ vs $6.08 \pm 1.23$) and understandable explanations ($\textbf{5.88} \pm \textbf{1.58}$ vs $5.29 \pm 1.72$) with $p < 0.01$ for both cases. For the \textbf{preference translator}, both models were able to extract relevant rules and generate acceptable explanations but no significant differences were observed.
Note, that this user study focused on evaluating the perceived quality, interpretability, and contextual relevance of the generated rules and preference vector explanations. The resulting preference vectors were not executed on the robot during the user study. Instead, the evaluation of robot behavior resulting from generated preference vectors was performed separately in Sec.~\ref{sec:quant} and \ref{sec:real}.

\subsection{Robot Experiments}
\label{sec:real}
\begin{figure}[t]
    \centering
    \begin{subfigure}{0.32\textwidth}
        \centering
        \includegraphics[width=\linewidth]{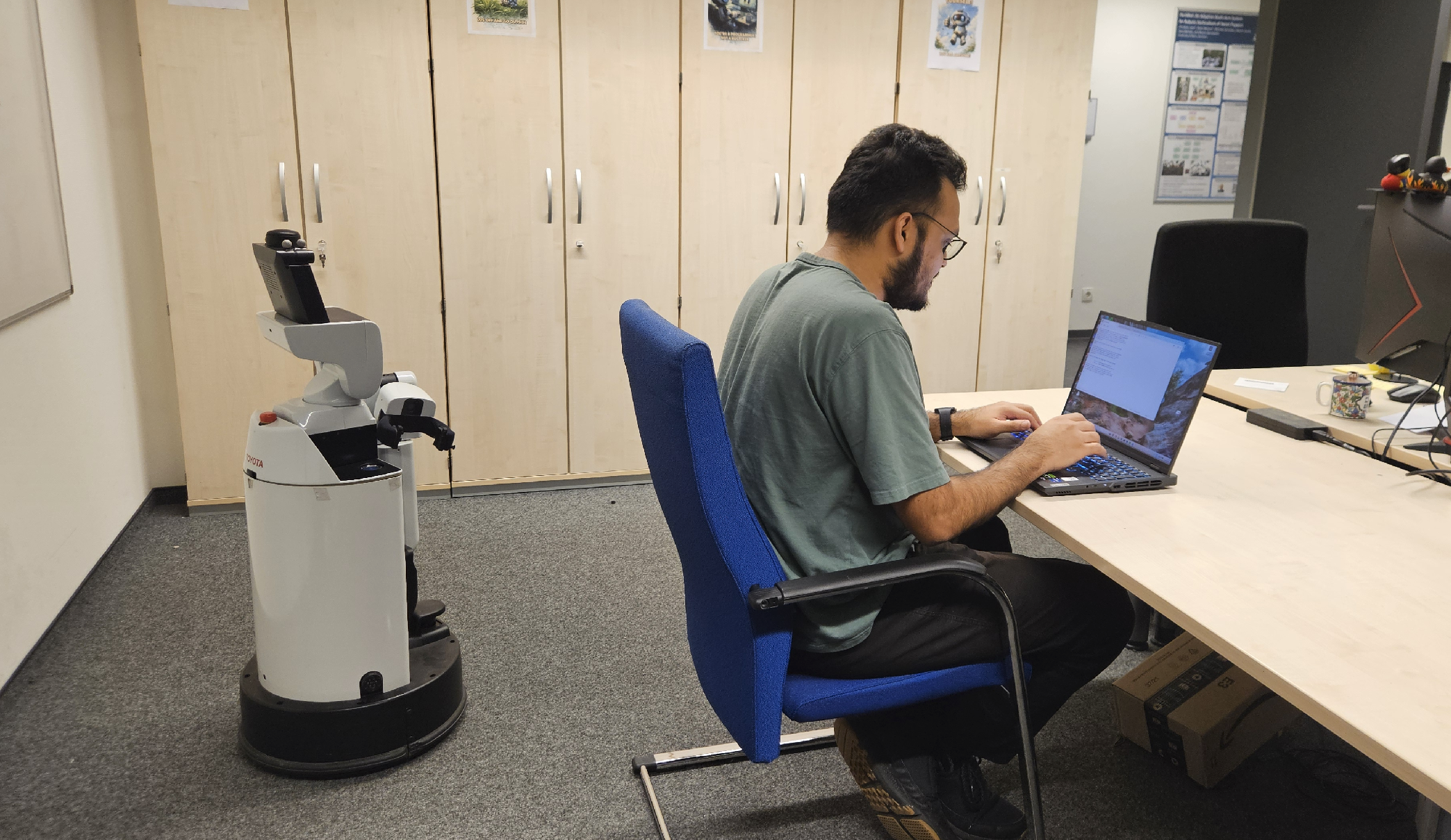}
        \caption{Office}
    \end{subfigure}
    \hfill
    \begin{subfigure}{0.32\textwidth}
        \centering
        \includegraphics[width=\linewidth, trim= 0 0 0 13, clip]{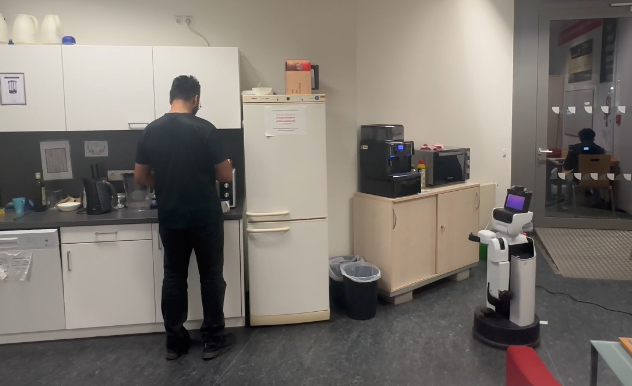}
        \caption{Kitchen}
    \end{subfigure}
    \hfill
    \begin{subfigure}{0.32\textwidth}
        \centering
        \includegraphics[width=\linewidth, trim= 0 0 0 13, clip]{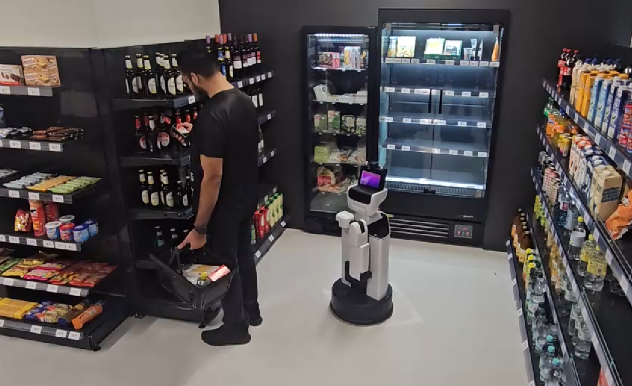}
        \caption{Supermarket}
    \end{subfigure}
    \caption{Real-world evaluation environments used for validating the proposed preference-aware navigation pipeline: (a) an office workspace, (b) a home kitchen environment, and (c) a supermarket grocery store. These environments were selected to capture varying spatial constraints, object densities, and human-interaction requirements.}
    \label{fig:real_world}
\end{figure}

\begin{table}[t]
\centering
\setlength{\tabcolsep}{3pt}
\textbf{Office}
\vspace{5px}

\begin{tabular}{l|c|c|c}
\hline
Group & Velocity (m/s) &  Human Dist. (m) & Traj. Length (m) \\
\hline
$\lambda_\mathit{{Base}}$ & $0.23 \pm 0.04$ & $1.01 \pm 0.09$ & $2.82 \pm 0.03$\\
$\lambda_\mathit{Hdist}$ & $0.17 \pm 0.01$ & $\textbf{1.25} \pm \textbf{0.03}$ & $2.66 \pm 0.08$\\
$\lambda_\mathit{velocity}$ & $\textbf{0.09} \pm \textbf{0.01}$ & $1.07 \pm 0.07$ & $2.83 \pm 0.03$\\
\hline
\end{tabular}

\vspace{5px}
\textbf{Home}
\vspace{5px}

\begin{tabular}{l|c|c|c}
\hline
Group & Velocity (m/s) &  Human Dist. (m) & Traj. Length (m)\\
\hline
$\lambda_\mathit{{Base}}$ & $0.24 \pm 0.02$ & $1.42 \pm 0.04$ & $3.26 \pm 0.10$\\
$\lambda_\mathit{Hdist}$ & $0.09 \pm 0.01$ & $\textbf{2.01} \pm \textbf{0.03}$ & $3.88 \pm 0.35$\\
$\lambda_\mathit{effic}$  & $0.09 \pm 0.02$ & $1.94 \pm 0.05$ & $\textbf{2.96} \pm \textbf{0.06}$\\
\hline
\end{tabular}

\vspace{5px}
\textbf{Supermarket}
\vspace{5px}

\begin{tabular}{l|c|c|c|c}
\hline
Group & Velocity (m/s) &   Human Dist. (m) &  Object Dist. (m)& Traj. Length (m) \\
\hline
$\lambda_\mathit{{Base}}$ & $0.20 \pm 0.02$  & $0.94 \pm 0.03$  & $0.94 \pm 0.03$ & $2.31 \pm 0.02$\\
$\lambda_\mathit{Hdist}$& $0.12 \pm 0.02$ & $\textbf{1.23} \pm \textbf{0.03}$ &- & $3.28 \pm 0.05$\\
$\lambda_\mathit{Odist}$ & $0.19 \pm 0.02$ & -& $\textbf{1.06} \pm \textbf{0.03}$ & $2.59 \pm 0.13$\\
\hline
\end{tabular}
\vspace{3px}
\caption{Results of the real robot experiments for preference-based navigation averaged over five runs. For each environment, two preference-vector-based navigation behaviors were evaluated. First, the preference to maximize distance from nearby humans ($\lambda_\mathit{Hdist}$) is activated in  all settings. Additionally, environment-specific preferences were applied: reducing velocity when navigating near humans ($\lambda_\mathit{velocity}$) in the office environment, maximizing navigation efficiency or shortest-path behavior ($\lambda_\mathit{effic}$) in the home (kitchen) environment, and maximizing distance from fragile glass bottles ($\lambda_\mathit{Odist}$) in the supermarket environment. Note that no human was present in the supermarket during the glass bottle experiments, and vice versa. Therefore, no values are reported for those cases. The results show consistent alignment between the activated preference dimensions and the resulting behavioral adaptations across environments.
}
\label{tab:all_groups_avg_pm}
\vspace{-20px}
\end{table}
 
To evaluate the real-world applicability and generalization capability of our proposed pipeline, we conducted a set of physical robot experiments using the Toyota HSR platform~\cite{yamaguchi2015hsr}.
Therefore, we evaluated the system in three representative indoor environments, i.e., an office workspace, a home kitchen, and a supermarket grocery store~(see Fig.~\ref{fig:real_world}). 
In each environment, we repeated a navigation experiment from fixed start and goal positions five times to assess robustness and repeatability.
The robot navigated between these positions while incorporating the natural language preferences specified for each experimental condition into its navigation strategy following our reasoning pipeline. 
During execution, we tracked the robot using a Vicon Vero motion capture system to compute average velocity, average distance to nearby humans and specific objects, and total trajectory length.
At each location, we first performed five baseline runs to establish the nominal performance of the RL agent without preference modulation.
This baseline served as a reference for evaluating the preference-driven behavioral adaptation.
Subsequently, we conducted two sets of preference-conditioned runs, one concerning distance to humans as a comparative case across all the environment, and the other selected randomly to exhaustively test the remaining preference vectors. 
In case of preference regarding distance to humans, the robot was instructed to maintain maximum distance from humans.
When a human was visible in the robot's camera, the pipeline would produce high values ($\geq 0.8$) for $\lambda_\mathit{Hdist}$ causing the robot to maintain larger distance between itself and the human while navigating towards the goal.
Similar adjustments occurred for the values of other components of the preference vector.

As shown in Table~\ref{tab:all_groups_avg_pm}, the MORL agent behavior consistently aligned with the provided context-dependent preferences across all evaluated environments. 
In particular, activating the human-distance preference ($\lambda_\mathit{Hdist}$) resulted in the largest average distance to humans in each scenario.
This consistent trend demonstrates that the generated preference vectors reliably modulate human-aware navigation behavior independent of the environment type.

Furthermore, the pipeline also successfully modulated task-specific navigation objectives. In the office environment, activating the velocity regulation preference ($\lambda_\mathit{velocity}$) resulted in the lowest average robot velocity, indicating effective motion adaptation in the vicinity of humans. 
In the home environment, prioritizing navigation efficiency ($\lambda_\mathit{effic}$) produced the shortest trajectory length, demonstrating improved path optimality while maintaining collision-free navigation.
Finally, in the supermarket environment, activating the object-distance preference ($\lambda_\mathit{Odist}$) increased the robot’s average distance to fragile objects compared to the baseline configuration. 
The relative improvement is smaller in this environment due to spatial constraints imposed by the store layout, which limit the feasible navigation space. 
Nevertheless, the observed behavior confirms that the MORL policy successfully incorporates object-aware safety preferences even in densely constrained environments.

Additionally, to assess the computational feasibility of our pipeline, we analyzed the latency of each module. It is important to note that the MORL policy executes continuously at high frequencies suitable for low-level control whereas the semantic pipeline operates asynchronously. The context predictor requires $\textbf{9.59} \pm \textbf{3.12 s} $ to generate the required context, whereas the rule updater takes about $\textbf{5.94} \pm \textbf{0.94 s} $ to update the existing ruleset. The translation of preference vector takes the least time at about $\textbf{3.34} \pm \textbf{0.89 s} $. Since this pipeline updates the preference vector asynchronously, no bottleneck occurs in the real-time execution of the MORL navigation policy. 

\subsection{Discussion}
Overall, these results demonstrate that the proposed pipeline enables consistent and controllable adaptation of robot navigation behavior according to context-dependent user preferences across diverse real-world environments.
Since the current framework builds on the multi-objective agent proposed by De Heuvel~\etal~\cite{morl-jorge-2025}, future work should investigate more general multi-objective formulations, such as model predictive control (MPC), to improve flexibility in dynamic settings.
In particular, MPC could be used to tune navigation parameters online by optimizing more diverse objectives over a receding horizon, such as safety, efficiency, comfort, social compliance, and user-specific preferences.
Furthermore, while the results are already promising, the robustness of our pipeline needs to be tested beyond the MIT Indoor Scenes and SRIN datasets.
We plan to evaluate our method in more diverse real-world locations to account for issues such as semantic ambiguity in highly cluttered scenarios which could lead to inconsistent identification and reasoning.
In particular, our pipeline currently may fail if the used foundation models hallucinate the presence or absence of humans or other important objects in the scene causing an unexpected change in the preference vector.
In such cases, a multi-model layer where different LLMs and VLMs rank and judge the generated reasoning and outputs could be beneficial.
By gathering and referencing interpretations across models, the system can achieve more stable reasoning and flag issues before they impact robot behavior.
Additionally, the performance might also be affected if the users provide conflicting preferences to the existing ruleset.
In such cases, it would be necessary to take feedback from the user regarding the update of the ruleset as opposed to letting the foundation model reason on its own.
Thus, our future work will include evaluations in more diverse and complex scenarios and also communication strategies between the robot and the user to shape trust, perceived control, predictability, and perceived acceptance. Additionally, we will investigate alternative low-level controllers to improve the robot's navigation behavior.


\section{Conclusion}
In this work, we presented a pipeline that enables robots to understand, represent, and apply context-dependent human navigation preferences by combining foundational models with a MORL policy. 
Our approach integrates high-level semantic reasoning with low-level motion control, allowing foundational models such as LLMs and VLMs to interpret scene context and user feedback while maintaining reliable real-time motion execution through the MORL policy. 
The proposed pipeline extracts structured environmental context, converts natural language preferences into interpretable behavioral rules, and translates these context-dependent rules into numerical preference vectors that directly modulate navigation behavior. 
Experimental results, including quantitative evaluations, a user study, and real-world robot deployments, demonstrate that the system can capture user intent and adapt navigation behavior across diverse environments. 



\begin{credits}
\subsubsection{\ackname} This work has partially been funded by the German Federal Ministry of Research, Technology and Space~(BMFTR) under the Robotics Institute Germany (RIG).

\subsubsection{\discintname}
The authors have no competing interests to declare that are relevant to the content of this article.
\end{credits}
%
%
%
%
\bibliographystyle{splncs04}
\bibliography{references}




\end{document}